\documentclass[10pt,twocolumn,letterpaper]{article}

\usepackage{wacv}
\usepackage{times}
\usepackage{epsfig}
\usepackage{graphicx}
\usepackage{amsmath}
\usepackage{amssymb}
\usepackage{balance}

\usepackage{multirow}
\usepackage{colortbl}
\def\ie{\emph{i.e.~}}

\def\etal{{\em et al.~}}

\wacvfinalcopy 

\begin{document}

\title{Digging Deeper into Egocentric Gaze Prediction}

\author{Hamed R. Tavakoli$^1$, Esa Rahtu$^2$, Juho Kannala$^1$, and Ali Borji\\
$^1$Department of Computer Science, Aalto University \\
$^2$Department of Signal Processing, Tampere University of Technology \\
{\tt\small 
hamed.r-tavakoli@aalto.fi, esa.rahtu@tut.fi, juho.kannala@aalto.fi, aliborji@gmail.com}
}

\maketitle
\ifwacvfinal\fi

\begin{abstract}
 This paper digs deeper into factors that influence egocentric gaze. Instead of training deep models for this purpose in a blind manner, we propose to inspect factors that contribute to gaze guidance during daily tasks. 
   Bottom-up saliency and optical flow are assessed versus strong spatial prior baselines. Task-specific cues such as vanishing point, manipulation point, and hand regions are analyzed as representatives of top-down information. 
   We also look into the contribution of these factors by investigating a simple recurrent neural model for ego-centric gaze prediction.
   First, deep features are extracted for all input video frames. 
   Then, a gated recurrent unit is employed to integrate information over time 
   and to predict the next fixation. We also propose
   an integrated model that combines the recurrent model with several top-down and
   bottom-up cues. 
   Extensive experiments over multiple datasets 
   reveal that (1) spatial biases are strong in egocentric videos,
   (2) bottom-up saliency models perform poorly in predicting gaze and underperform spatial biases, 
   (3) deep features perform better compared to traditional features, 
   (4) as opposed to hand regions, the manipulation point is a strong influential cue for gaze
   prediction, 
   (5) combining the proposed recurrent model with bottom-up cues, vanishing points and, in particular, 
   manipulation point results in the best gaze prediction accuracy over egocentric videos,
   (6) the knowledge transfer works best for cases where the tasks or sequences are similar,
   and
   (7) task and activity recognition can benefit from gaze prediction. Our findings suggest that (1) there should be more emphasis 
   on hand-object interaction
    and (2) the egocentric vision community should consider 
    larger datasets including diverse stimuli and more subjects.

\end{abstract}

\section{Introduction}

Gaze prediction
  in egocentric (first person) vision, contrary to traditional gaze prediction in free-viewing setups,
 is an unconstrained challenging problem in which many factors including 
 bottom-up saliency information, task specific dependencies, individual subject
 variables (e.g., fatigue, stress, interest) contribute.
This paper evaluates various components of visual attention, including top-down
 and bottom-up elements, that may contribute to the prediction of
 gaze location in egocentric videos.


Egocentric vision considers the analysis of the visual content of daily activities from 
mass-marketed miniaturized wearable cameras such as cell phones, GoPro cameras,
 and Google glass. Analysis of images captured by egocentric cameras can 
 reveal a lot about the person recording such images, including, intentions, personality, interests, etc. 
First-person gaze prediction is useful in a wide range of applications in health care,
 education and entertainment, for tasks such as action and event recognition~\cite{pirsiavash2012detecting}, 
 recognition of handled objects~\cite{ren2009egocentric}, discovering important 
 people~\cite{ghosh2012discovering}, video re-editing~\cite{jain2015gaze}, 
 video summarization~\cite{Xu_2015_CVPR}, engagement detection~\cite{su2016detecting}, 
 and assistive vision systems~\cite{hadsell2006dimensionality}. 


To date, in computer vision community, the study of gaze behavior 
has been mainly focused on understanding free-viewing gaze guidance.
Consequently, while it is possible to accurately measure the gap between human inter-observer model and 
computational models in this task~\cite{borji2013state}, our understanding
of top-down and task driven gaze, despite its prevalence in daily vision~\cite{land2001ways}, is relatively
limited. 
 
This paper explores bottom-up and top-down attentional cues involved in guiding first person gaze guidance. The role of spatial biases are studied using a central Gaussian map, the average fixation map, and a fixation oracle model. Further, the contribution of bottom-up saliency, vanishing point, and optical flow are investigated. as representatives of bottom-up cues. The studied task specific cues include manipulation point and hand regions. A deep model of gaze prediction is also developed as a proxy to deep models such as~\cite{Zhang2017} integrating multiple cues implicitly.
A set of extensive experiments is conducted to determine the contribution of each factor as well as their combination.


\section{Related Works}

\noindent With the increasing popularity of egocentric vision in recent years, numerous research work is being focused to solve the computer vision 
problems from the first person perspective.
First person vision problems are unlike classic computer vision problems since the person whose actions are being recorded is not captured. Egocentric vision poses unique challenges like non-static cameras, unusual viewpoints, motion blur, variations in illumination with the varying positions of camera wearer, real time video analysis requirements, etcetera. Tan \etal~\cite{fpvvstpv} demonstrate that challenges posed by egocentric vision can be handled in a more efficient manner if analyzed differently than exocentric vision. 
Substantial research has tried to address various computer vision problems such as object understanding, object detection, tracking, and activity recognition, from the egocentric perspective.
We refer the readers to~\cite{betancourt2015evolution} for a recent review on the applications 
of first person vision.

\subsection{Attention in Egocentric Vision}
Yamada \etal~\cite{yamada2010can} found that conventional saliency maps can predict egocentric fixations better than chance and that the accuracy decreases significantly with an increase in ego-motion. 
Matsuo \etal\cite{matsuo2014attention} proposed to combine motion and visual saliency to predict egocentric gaze. Park \etal~\cite{park20123d} introduced a model to compute social saliency from head-mounted cameras to recognize gaze concurrences. Li \etal~\cite{li2013learning} proposed to predict gaze in egocentric activities involving meal preparation by combining implicit cues from visual features such as hand location and pose as well as head and hand motion (see also~\cite{nakashima2015saliency}). Camera motion has been
shown to represent a strong cue for gaze prediction in~\cite{matsuo2014attention}.
Polatsek \etal~\cite{polatseknovelty} present a model based on spatiotemporal visual information captured from the wearer's camera, specifically extended using a subjective
function of surprise by means of motion memory. Su and Grauman~\cite{su2016detecting} proposed a learning-based approach that uses long-term egomotion cues to detect user engagement during an activity (e.g., shopping, touring). 
Yonetani \etal~\cite{yonetani2016visual} proposed a method to discover visual motifs, images of visual experiences that are significant
and shared across many people, from a collection of first-person videos. Bertasius~\etal~\cite{bertasius2016first} proposed the \textit{EgoNet} network and the idea of \textit{action-objects} to approximate momentary visual attention and motor action with objects, without gaze tracking or tactile sensors. Zhang \etal~\cite{Zhang2017} proposed training a model for predicting the gaze on future frames.
They initially generate several future frames given a latent representation of the current frame using an adversarial architecture as in video generation techniques~\cite{vondrick2016nips}.
Then, a 3D convolutional neural network is employed on the generated frames to estimate the 
gaze for the $50$-th frame from the current frame. Recently, Huang \etal~\cite{huang2018predicting} proposed 
a hybrid model based on deep neural networks to integrate task-dependent attention transition with bottom-up saliency. Notice that here we do not intend to benchmark these models, rather our main goal is to understand factors that influence gaze in daily tasks.

\subsection{Top-down Visual Attention and Video Saliency}
Navalpakkam and Itti~\cite{navalpakkam2006integrated} proposed a cognitive framework for task-driven attention using four components: 1) determining task-relevance of an entity, 2) biasing attention towards target features, 3) recognizing targets using the same low-level features, and 4) incrementally building a task-relevance map. Some models have incorporated Bayesian and reinforcement learning techniques including (e.g.,~\cite{sprague2003eye}). Peters and Itti~\cite{peters2007beyond} and Borji \etal~\cite{borji2012probabilistic} used classification techniques such as Regression, SVM, and kNN to map a scene gist, extracted from the intermediate channels of the Itti saliency model~\cite{itti1998model}, to fixations.

Some studies have investigated eye-hand coordination during grasp and object manipulation. For example, \cite{belardinelli2016s} studied the bidirectional sensori-motor coupling of eye-hand
coordination. In a task where subjects were asked to either pretend to drink out of the
presented object or to hand it over to the experimenter, they found that fixations show a
clear anticipatory preference for the region where the index finger is going to be placed. Task-driven attention has also been studied in the context of joint attention during childhood as an important cue for learning (e.g.,~\cite{yu2013joint}). In addition, several works have studied gaze guidance during natural behavior in tasks such as sandwich making, walking, playing cricket, playing billiard, and drawing.

A tremendous amount of research has been conducted on predicting fixations over still images and videos (See~\cite{borji2013state} for a review). Traditionally, spatial saliency models have been extended to the video domain by adding a motion channel. Some models have computed video saliency in the frequency domain (e.g.,~\cite{guo2008spatio}). 
Seo and Milanfar~\cite{seo2009static} utilized self similarities of spatio-temporal volumes to predict saliency. Itti and Baldi defined video saliency as Bayesian Surprise~\cite{itti2005bayesian}.
Rudoy \etal~\cite{rudoy2013learning} proposed a learning-based framework for saliency prediction. A number of recent models have utilized deep learning for this purpose (e.g.,~\cite{bazzani2016recurrent}). For instance, Cagdas \etal~\cite{bak2016two}, inspired by~\cite{simonyan2014two}, proposed a two stream CNN for video saliency, one stream built on appearance and another on motion.

\section{Methods}
Our approach to understand the egocentric gaze guidance consists of two steps, (1) 
model-free evaluation to assess contribution of each cue separately and in conjunction with other cues, and 2) a model-based analysis by building computational models. 
For each cue, a specific computational model is developed. The computational methods are based on (1) regression from feature domain to saliency domain, (2) traditional bottom-up saliency prediction models,
and (3) deep learning models. We will discuss the details of regression and deep models of gaze prediction next.
Details of feature cues will be discussed in section~\ref{sec:analysis}.

\subsection{Regression}
Here, the ground-truth fixation map is initially smoothed in order to reduce (a) the randomness of landing eye movements by viewers, and (b) eye tracking error. Then, a regressor from the feature space to fixations is learned. Assume each frame is encoded by a feature vector of size $1\times m$. Vectors of all $n$ frames are vertically stacked leading to the $n \times m$ matrix $M$. Each ground-truth fixation map has one at the location of the gaze and zeros, elsewhere (over a $k \times k$ grid map, here $k$=20). This map is first convolved with a small isotropic Gaussian (width 5, sigma 1) and is then linearized. By vertically stacking these vectors over all $n$ frames (as above) we will have the matrix $X$ of size $n \times k^2$. Our goal is to find vector $W$ (of size $m \times k^2$) to minimize $||M×W - X ||^2_{2}$. This is a least square problem and can be solved through SVD decomposition as,
\begin{equation}
M \times W = X, \  W = M^+ \times X 
\end{equation}
where $M^{+}$ is the pseudo-inverse of matrix $M$ (i.e., $(M^{T}M)^{-1}M^{T}$. 
For a test frame, we first extract feature map $F$ and then generate the prediction map as $P = F \times W$ which is then reshaped to a $k \times k$ gaze probability map. 

\subsection{Deep Models}

\paragraph{Deep Regression} 
To investigate the power of features obtained from CNNs, we learn a regressor from
frames encoded using 3 architectures, namely: Inception, ResNet and VGG16.
It is worth noting that such features also encode the global context of a frame.

\paragraph{Gated Recurrent Units (GRUs)}
A deficiency of deep regression model is overlooking temporal information. Such information can be utilized using recurrent models. 
The input egocentric video frames are fed to a pretrained CNN (over ImageNet~\cite{deng2009imagenet}) and then extracted features from different layers are used to train a GRU~\cite{chung2014empirical} to predict fixations.
The task of gaze prediction at time T is to estimate the following probability,

\begin{equation}
    p(g_1, \ldots, g_{T}, x_1, \ldots, x_T) = 
   \prod_{t=1}^T p(g_t| g_{t<T}, x_{\leq T})
    \label{eq:prob}
\end{equation}
\noindent where $g_i$ and $x_i$ are the 2D gaze location and feature representation of the $i-$th frame in the video, respectively. Given previous fixations and frames, the goal is to predict the gaze location over the current time $T$. Here, we assume that previous fixation data is not available so the goal is to predict the fixation location given only video frames up to the current time (\ie, offline case similar to Li~\etal~\cite{li2013learning}). In this case, the above joint probability distribution reduces to $\prod_{t=1}^T p(G_t| x_{\leq T})$. 
We utilize a GRU architecture with the following formulation:
\begin{eqnarray}
    z_t &=& \sigma_g(W_{z} x_t + U_{z} h_{t-1} + b_z) \\
    r_t &=& \sigma_g(W_{r} x_t + U_{r} h_{t-1} + b_r)      \\ 
    h_t &=&  z_t \circ h_{t-1} + \\
    && (1-z_t) \circ \sigma_h(W_{h} x_t + U_{h} (r_t \circ h_{t-1}) + b_h)
\end{eqnarray}
    
\noindent where $x_t$ is the input, $h_t$ is the output, $z_t$ and $r$ are the update and reset gates, respectively. $W$, $U$ and $b$ are the weights and bias to be learned. $\sigma_g$ and $\sigma_h$ are sigmoid and hyperbolic tangent functions, respectively.

Each video frame (RGB images with resolution $ 640 \times 480 $) is first fed to a pretrained CNN and a feature vector $x$ is extracted from either the fully connected layers or the final class label layer.
The corresponding 2D gaze vector for each video frame is extracted and converted to a $20 \times 20$ sparse binary map with a 1 at the location of gaze and 0, elsewhere. This map is then linearized to a 400D vector and is used as the output vector in training GRU.

\paragraph{Network Training} 
The proposed architecture consists of three stacked GRU units. 
Each GRU has 20 hidden states, and has a step size of 6. We implemented the architecture with tensorflow 
~\cite{tensorflow2015-whitepaper}. We train the model using cross entropy loss with softmax activation functions to discriminate between fixated locations and non-fixated ones. That is,
\begin{equation}
    \mathcal{L}(y,x) = -\sum_i y^{(i)}\log(p(\hat{y}|x)),
\end{equation}
\noindent where $x$ is the input feature, $y$ is the ground-truth indicating if there exists a fixation or not, and $\hat{y}$, is associated with the predicted fixation. We train the model to predict which of the 400 possible locations is fixated. 
In other words, during the training, the ground truth data is treated as a one-hot vector defining which location is fixated at each frame. We then employed Adam optimizer~\cite{kingma2014adam} with learning rate of 0.0001 for 25 epochs. Although we follow a classification scheme to form a saliency map, we adopt a regression interpretation of the output of the model.

\section{Data and Evaluation Criteria}
\subsection{Datasets} We utilize 3 datasets. The sequences consist of video game playing~\cite{borji2012probabilistic}, cooking and meal preparation~\cite{fathi2012learning} tasks. Table~\ref{tab:DataStat} shows summary statistics of these datasets.

\begin{table}[hbt!]
\caption{Summary statistics of the utilized datasets. USC videos are cartoonic outdoor while GTEA videos are natural and indoor. Hands are often visible in GTEA but not over USC videos. }

\label{tab:DataStat}
\begin{center}\mbox{
\begin{footnotesize}
\begin{tabular}{l|llllc}
\renewcommand{\arraystretch}{.2}
\renewcommand{\tabcolsep}{.7mm}

 \multirow{2}{*}{Dataset} &  Game/ & Frames &   Avg. Video & Size & Hands    \\
 &  Task & & Duration (min)& MB  &  Visible?   \\
\hline
\hline

 \multirow{2}{*}{USC}  &  3DDS & 90000  & 11.59 $\pm$ 0 &   433 & N  \\

 & HDB & 45000  & 5.59 $\pm$ 0  & 216  & '' \\

\hline

 \multirow{2}{*}{GTEA}  & Pizza & 75161 & 10.18 $\pm$ 2.22& 3630 & Y  \\

 & Snack & 69775 & 7.68 $\pm$ 0.76 & 2741  & '' \\

Gaze+ & American & 96297 & 13.06 $\pm$ 1.01 & 4651 & '' \\

\hline

GTEA   & Sandwich  &  \multirow{2}{*}{35730} &   \multirow{2}{*}{3.21 $\pm$ 1.28} &   \multirow{2}{*}{391} & \multirow{2}{*}{ Y} \\
 Gaze &  making &  &   &   &  


\end{tabular}
\end{footnotesize}
}\end{center}
\end{table}

 \paragraph{USC Video Games} 
 Data, including frames, fixations, and motor actions were collected by Borji \etal~\cite{borji2012probabilistic} using an infrared eye tracker while subjects played video games. We use data of two games. The first one called 
\textit{3D Driving School (3DDS)} is a driving emulator with simulated
traffic conditions. An instructor will
tell the players the direction in a semi-translucent
text box above the screen and/or a small arrow on the top-left
corner. The second game called \textit{Hot Dog Bush (HDB)} is a 2D time management game. Players are supposed to serve customers hot-dogs by assembling ingredients placed at different stations. 
Later in the game, customers can also order drinks. Players should trash burned sausages and collect the payments.

\paragraph{GTEA Gaze/Gaze+ Datasets} 
We also utilize two datasets collected by Fathi \etal~\cite{fathi2012learning}. The first one is \textit{GTEA Gaze+} dataset. We chose a subset of this dataset (as in~\cite{li2013learning}; first 15 videos of 5 subjects and 3 recipes including Pizza, American breakfast, and Afternoon snack). 
We report accuracies over each recipe. The second dataset, known as \textit{GTEA Gaze}, includes 17 sequences
performed by 14 different subjects. Both datasets contain videos of meal preparation (the first one includes sandwich making from ingredients on a table and the second one is cooking in a kitchen) with head-mounted gaze tracking and action annotations. All videos involve sequential object interaction.

\subsection{Evaluation Criteria} We utilize two scores, Area Under the Curve (AUC) and Normalized Scanpath Saliency (NSS), 
to measure the consistency between observers' fixations and models' predictions. Please refer to~\cite{borji2013quantitative} for detailed definitions.

\section{Analysis}
\label{sec:analysis}

\begin{figure}[t]
\begin{center}
\includegraphics[width=8.2cm,height=6cm]{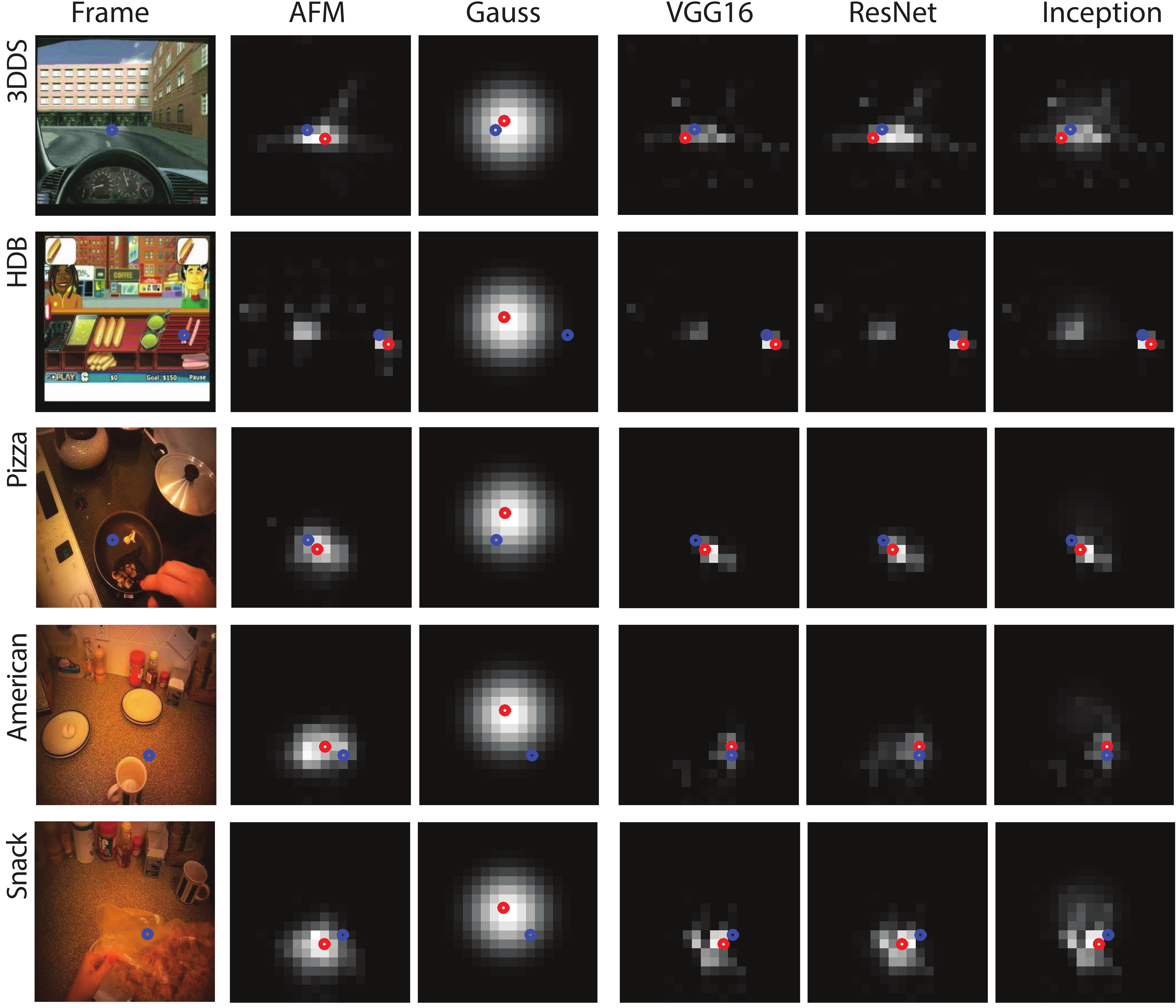}
\end{center}
\caption{Sample frames of video games and egocentric videos along with predictions of recurrent deep model using various CNNs. Blue and red circles denote the ground truth fixation location and maximum of each map, respectively. Notice a sharp bias towards the bottom part of the scene over the GTEA dataset induced by table-top objects and hands.}
\label{fig:MapSamples}
\end{figure}

\subsection{Spatial Biases} 
The spatial bias is a strong baseline in predicting the fixation locations~\cite{tatler2007central}.
To investigate its role in egocentric vision, we employed three baselines (1) \textbf{Central Gaussian Map (Gauss)}, (2) \textbf{Average Fixation Map (AFM)}, and (3) \textbf{Fixation Oracle Model (FOM)}. 
The central Gaussian model is motivated by the human tendency to look at the center of images in free-viewing. 
The AFM is the average of all training fixations, which forms a spatial prior. The FOM is the upper-bound, obtained by convolving the ground-truth fixation maps with a Gaussian kernel. Fig.~\ref{fig:MapSamples} visualizes sample video frames, the spatial biases and example predictions with
the actual gaze point overlaid.

Table~\ref{tab:SP} summarizes the performance of spatial bias models. 
The AFM model is the best model and even exceeds the performance of~\cite{li2013learning}. 
This indicates a strong central bias in egocentric gaze estimation and is alerting for the current computational models of gaze estimation 
as they can be replaced by a simple Gaussian model, learned from the same sequences of data used for training such models. 
The FOM model scores AUC of $0.97$ and NSS of $19.95$ over two datasets (same on each video). 
This suggests that there is still a large gap between existing models and human performance in predicting egocentric gaze.

\begin{table}[t]
\caption{The performance of spatial biases in comparison to bottom-up models and Li~\etal~\cite{li2013learning}. The 1st row is NSS and the 2nd row is AUC. The AFM outperforms All-BU, a mixture of bottom-up models and Li~\etal~\cite{li2013learning}. }
\label{tab:SP}
\begin{center}
\begin{small}
\begin{tabular}{l||c|c|c|c}
\renewcommand{\tabcolsep}{.3mm}
\renewcommand\arraystretch{.7}
{Video}  & {AFM} & {Gauss}  & All-BU & {Li~\etal~\cite{li2013learning}} \\
    \hline
	\hline
 \multirow{2}{*}{3DDS}  &  1.588 & \textbf{1.705}& 1.112 & \multirow{4}{*}{-}  \\
    & 0.796  &\textbf{0.814} & 0.768 & \\

 \multirow{ 2}{*}{HDB}  &   \textbf{2.052}   &0.731 &  0.902 &\\
		                 & \textbf{0.740}  &  0.700& 0.675 &  \\
\hline
 \multirow{ 2}{*}{Pizza}  &  \textbf{1.682} & 1.467 & 1.064& \multirow{6}{*}{AUC = 0.867}\\
							                         & \textbf{0.885}  &   0.829 & 0.767 & \\
 \multirow{ 2}{*}{Snack}  &   \textbf{2.040}   & 1.076 &  1.176 &  \\
    & \textbf{0.893}  & 0.784  &  0.775 &\\

 \multirow{ 2}{*}{American}  &   \textbf{1.986} & 1.164 & 1.101 & \\
                              &  \textbf{0.888}  & 0.803  & 0.774 & \end{tabular}
\end{small}     
\end{center}
\end{table}

\subsection{Bottom-up Saliency and Optical Flow}
We computed the maps from 3 classic saliency models including \textit{Itti}~\cite{itti1998model}, \textit{GBVS}~\cite{harel2006graph}, and \textit{SR}~\cite{hou2007saliency}. We, then, used the saliency maps for predicting the gaze. The saliency maps were further complemented with motion features as an important source of information that influences the attention in videos. For motion features, we computed the optical
flow (OF) magnitude using the Horn-Schunck algorithm~\cite{horn1981determining} as
a cue that captures both ego-motion and global motion.
The optical flow magnitude map was then employed for predicting egocentric fixations.

We also combined the saliency maps from saliency models and optical flow to predict gaze.
To this end, we trained a regressor that combines the feature maps from the three saliency 
models and the motion features. We will refer to this model as "all bottom-up" (All-BU) model 
in the rest of this paper as the representative of the bottom-up features.

\begin{table}[t]

\caption{Gaze prediction accuracy of BU models and their combination (1st is NSS). OF stands for optical flow magnitude. Top winner is shown in bold. GBVS does better than other models (except combination of all) due to its smoother maps.}
\label{tab:BU}
\begin{center}
\renewcommand\arraystretch{.8}
\renewcommand{\tabcolsep}{2.7mm}
\begin{small}
\begin{tabular}{l||c|ccc|c}

  Video & OF & Itti & GBVS & SR  & All-BU\\
  \hline
    \hline
 \multirow{ 2}{*}{3DDS} & 0.229 & 0.834 & 1.067 & 0.160 & \textbf{1.112} \\
                                   & 0.623 & 0.723 & 0.760 & 0.572  & \textbf{0.768}\\
 
 \cline{2-6}

 \multirow{ 2}{*}{HDB} & \textbf{0.981} & 0.235 & 0.722 & 0.334  & 0.902\\
& 0.618 &  0.581  & 0.667 &  0.541 &  \textbf{0.675}  \\
  
  \hline

 \multirow{ 2}{*}{Pizza}  & 0.479 & 0.806 &1.064 & 0.774 & \textbf{1.064}\\
 & 0.675 & 0.719 & 0.764 &0.740 &  \textbf{0.767}\\
 \cline{2-6}

 \multirow{ 2}{*}{Snack} & 0.486 & 0.872 & 1.157 & 0.816 & \textbf{1.176}\\
 &   0.689  & 0.734 &  0.772 & 0.722 &  \textbf{0.775} \\
 \cline{2-6}
 
 \multirow{ 2}{*}{American} & 0.454 &  0.875 & 1.086 & 0.607  & \textbf{1.101}\\
 &  0.702  & 0.737 &  0.774 & 0.711 & \textbf{0.774} \\

\end{tabular}
\end{small}
\end{center}
\end{table}

Table~\ref{tab:BU} shows the results of saliency models, optical flow magnitude, and the combination of 
all of them. There is no single model that outperforms the combination of all models. 
The optical flow is not also a strong predictor alone except for sequences
with highly moving objects like HDB (in HDB the scene is static, but it includes several moving objects).
Nevertheless, the combination of optical flow and all other 
saliency models improves the results. 
This indicates that BU saliency models fail to capture egocentric gaze.

Table~\ref{tab:SP} shows that BU models underperform spatial biases. Even a state of the art saliency model known as SALICON~\cite{huang2015salicon}, 
did not perform well on this data. It achieves NSS of $0.98$ and $0.81$ over 3DDS and HDB games (almost as good as GBVS). These results indicate that low level saliency only weakly contributes to egocentric gaze.

\subsection{Deep Models}
The performance of deep models is summarized in Table~\ref{tab:deep}.
The GRU model with inception features outperforms all other models over all videos in terms of AUC, except for Snack video where Resent features perform better.

 \begin{figure}[t]
  \centering
    \includegraphics[width=8.2cm,height=3.3cm]{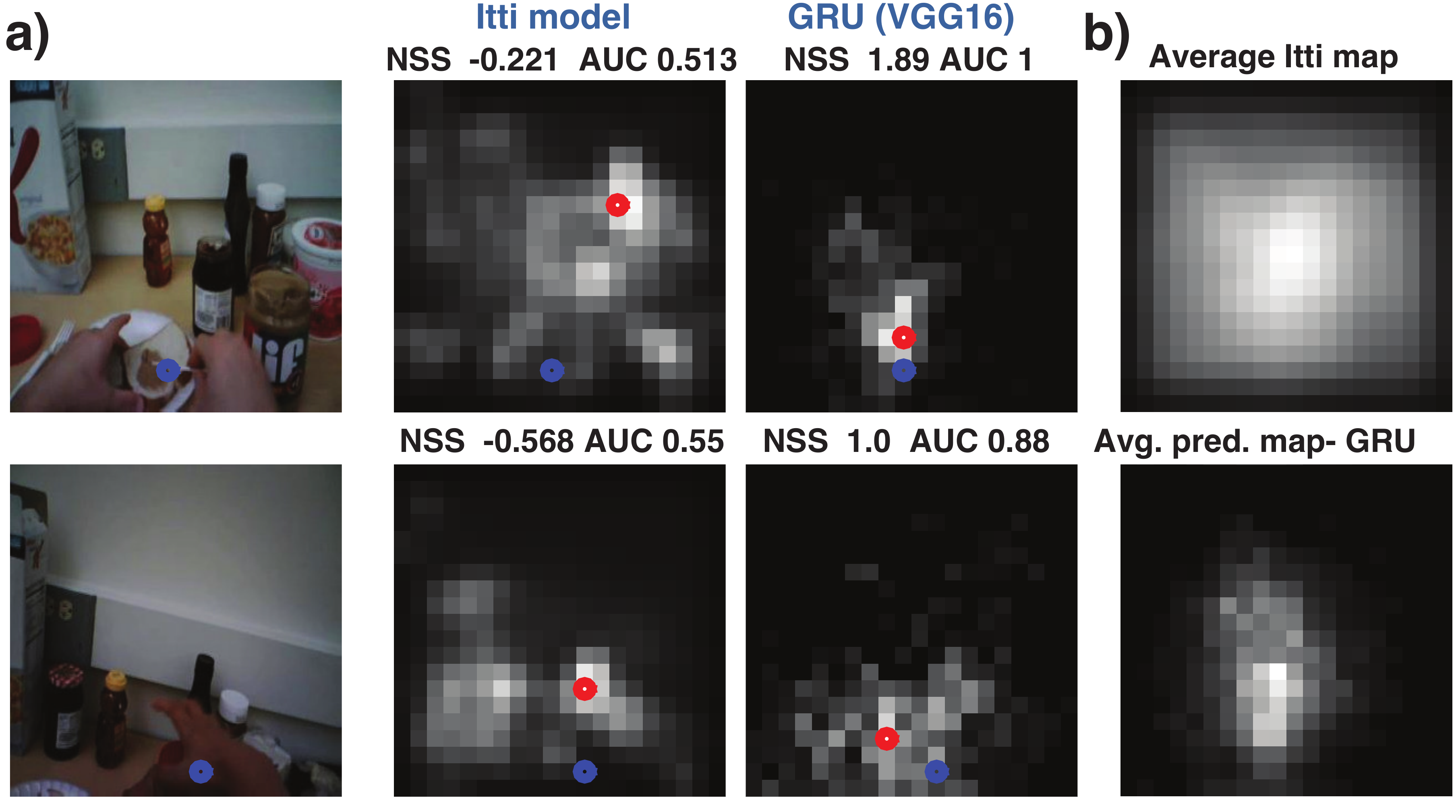}
  \caption{a) Two sample frames from the GTEA Gaze dataset along with maps from Itti and the recurrent deep model (indicated GRU), b) Average bottom-up saliency and average prediction map of our model using VGG16 features. Our model generates more focused maps. Blue and red circles denote gaze location and map maximum, respectively.}
  \label{fig:BU}
  \end{figure}

 \begin{table}[b]
 \caption{Performance of deep models. NSS (1st row) and  AUC (2nd row) scores of regression and recurrent model. For the ResNet and Inception CNNs, we use the class layer probabilities (a 1000D vector) to represent the video frames while for VGG16, we use the output of the last fully connected layer. The best accuracy in each row is shown in \textbf{bold}.}
\begin{center}
\renewcommand{\tabcolsep}{0.6mm}
\renewcommand\arraystretch{.8}
\begin{small}
\begin{tabular}{l||ccc|ccc}
  \multirow{3}{*}{Video}  & \multicolumn{3}{c|}{Regression} & \multicolumn{3}{c}{Recurrent}  \\
\cline{2-7}
    & VGG16 &  ResNet  & Inception  &  VGG16 &  ResNet  & Inception  \\
 &  2048D  & 1000D & 1000D  &  2048D  & 1000D & 1000D  \\
    \hline
        \hline
 \multirow{ 2}{*}{3DDS}  & 1.498 & \textbf{1.588} & \textbf{1.588} & 1.317 &  1.548  & 1.530  \\
				     &  0.805 &   0.797 &  0.796  &0.752 &  0.810  & \textbf{0.815}  \\
\cline{2-7}

 \multirow{ 2}{*}{HDB}  &  \textbf{2.665} & 2.129 & 2.111 & 1.692 &  1.748  & 1.798  \\
				   &  0.790 & 0.746   & 0.752   & 0.800 &  0.807  & \textbf{0.822}  \\
\hline

 \multirow{ 2}{*}{Pizza}  & 1.387 & 1.720 &  1.640 & 1.650 &  \textbf{1.748}  & 1.696  \\
 &  0.833 &   0.875 &  0.869  & 0.842 &  0.857  &  \textbf{0.877}  \\
\cline{2-7}
 \multirow{ 2}{*}{Snack}  &  1.759 & \textbf{2.011} & 1.992 & 1.604 &  1.827  & 1.709 \\
 & 0.879 & 0.882   & 0.882   & 0.833 &  \textbf{0.865}  & 0.864 \\
\cline{2-7}
 \multirow{ 2}{*}{American}  &  1.412 & \textbf{2.045} & 1.884 & 1.702 &  1.717  & 1.984\\
 &  0.868 & 0.881   & 0.868   & 0.837 &  0.868  & \textbf{0.890}\\
\end{tabular}
\end{small}
\end{center}
\label{tab:deep}
\end{table}

Compared to the spatial biases, in particular the AFM, deep features perform better for both architectures.
The deep regression model scores well in terms of NSS. 
In terms of AUC score, however, the recurrent model performs the best. 
NSS score is a crude measure. To dig dipper into results, 
we calculated the percentage of video frames for which our model produces positive NSS scores.
We found that on average about 75\% of the frames have NSS values above zero, and approximately 69\% of the frames have NSS values above one. There are only 25\% of frames with negative NSS. Considering both scores, 
this analysis indicates that the recurrent model approximates the fixated regions better than baselines.

The deep models not only have a better performance in comparison to~\cite{li2013learning}, but also are much simpler as they do not need hand detection or head motion estimation. 
The model by Li \etal~\cite{li2013learning} scores an average AUC of $0.867$ over the GTEA gaze+ dataset which is below $0.877$, the best average AUC of deep models. 
This somewhat indicates that deep models may implicitly capture some top-down factors.
Fig.~\ref{fig:BU}.a shows sample frames from the GTEA gaze dataset along with their corresponding maps from Itti and deep models.
Itti model scores AUC of $0.749$ and NSS of $1.064$ on this dataset. Contrary to bottom-up saliency models, our deep models successfully highlight task-relevant regions. For example, as depicted in Fig.~\ref{fig:BU}.b, the deep recurrent model predicts the gazed regions more effectively.

\subsection{Task-specific cues}
\label{taskspecific}
Here, we look into the utility of task-specific factors, including  (a) \textit{vanishing point}, (b) \textit{manipulation point}, and (c) \textit{hand regions}. 
Certain cognitive factors are believed to guide attention in specific tasks~\cite{land2001ways}. 
For example, drivers pay close attention to the road tangent or vanishing point while driving~\cite{land2001ways}.
During making coffee attention is sequentially allocated to task-relevant entities such as coffee mug, coffee machine, and object manipulation points because hands are tightly related to objects in manipulation, reaching and grasping~\cite{belardinelli2016s}. 
\begin{figure}[t]
  \centering
    \includegraphics[width=8.3cm,height=10cm]{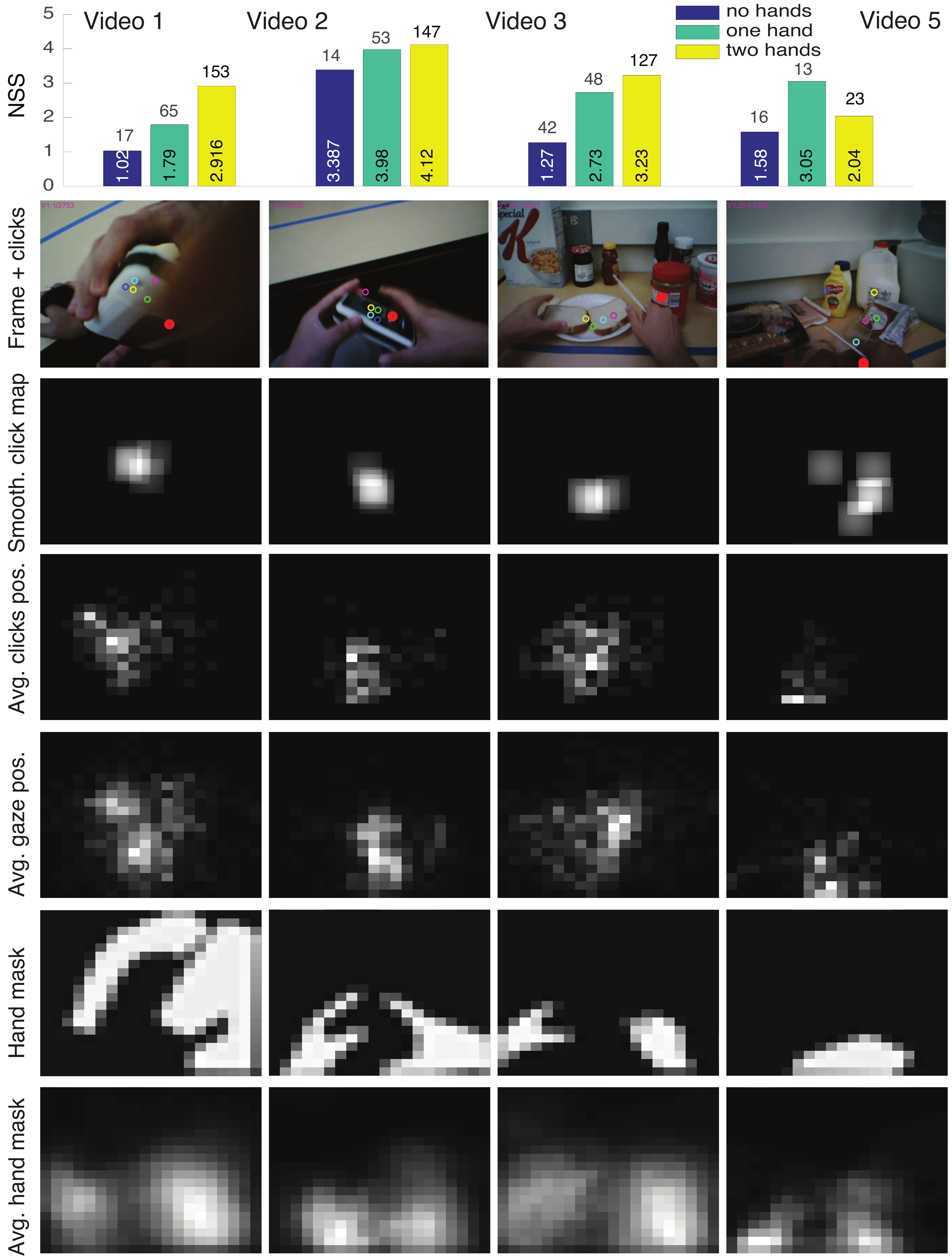}
  \caption{The role of manipulation point and hand regions on fixation prediction. Rows from top to bottom: \textbf{First:} Results of the behavioral experiment showing NSS scores for frames with no-hand, one hand and two-hands for first four videos of GTEA Gaze dataset. \textbf{Second:} Sample frames along with clicked locations by five subjects. \textbf{Third:} Smoothed click maps. \textbf{Fourth:} Average click locations over each video. \textbf{Fifth:} Ground-truth fixation locations over all frames of each video. \textbf{Sixth:} Annotated hand regions. \textbf{Seventh:} Average hand masks for each video.}
  \label{fig:clicks}
\end{figure}
\paragraph{Vanishing points (VP)}
To assess whether and how much this cue can improve accuracy of gaze prediction, we ran the vanishing point detection algorithm of~\cite{borji2016vanishing} on 3DDS frames. We chose 3DDS as it is a driving game task with vanishing points in the sequence. 
This algorithm outputs a $20 \times 20$ binary map with 1 at the VP location and zeros, elsewhere. We then convolved this map with a small Gaussian kernel to obtain a vanishing point map. 
This map scores AUC of $0.763$ and NSS of $1.443$ which are much higher than chance but still below spatial biases. 
\paragraph{Hand regions} 
To investigate whether hands predict gaze, we manually annotated hand regions over first $4$ videos of the GTEA gaze dataset as depicted in Fig.~\ref{fig:clicks}, $6$-th row. Then, we employed the binary hand mask to predict fixations in the frames with hands.
NSS scores over 4 videos in order are: $-0.37, -0.28, -0.31,$ and $-0.22$. 
The negative NSS values indicate that the hand masks predict regions with no fixation because
NSS(1-S) = - NSS(S) where 1-S is the complement of map S. 
Thus, fixations often fall outside hand regions which means that 
their complement map is predictive of fixations. 
This indicates that hands by themselves are not informative of gaze,
rather their presence in conjunction with the manipulated object is useful. 

\begin{figure}[t!]
  \centering
    \includegraphics[width=8.3cm,height=3.7cm]{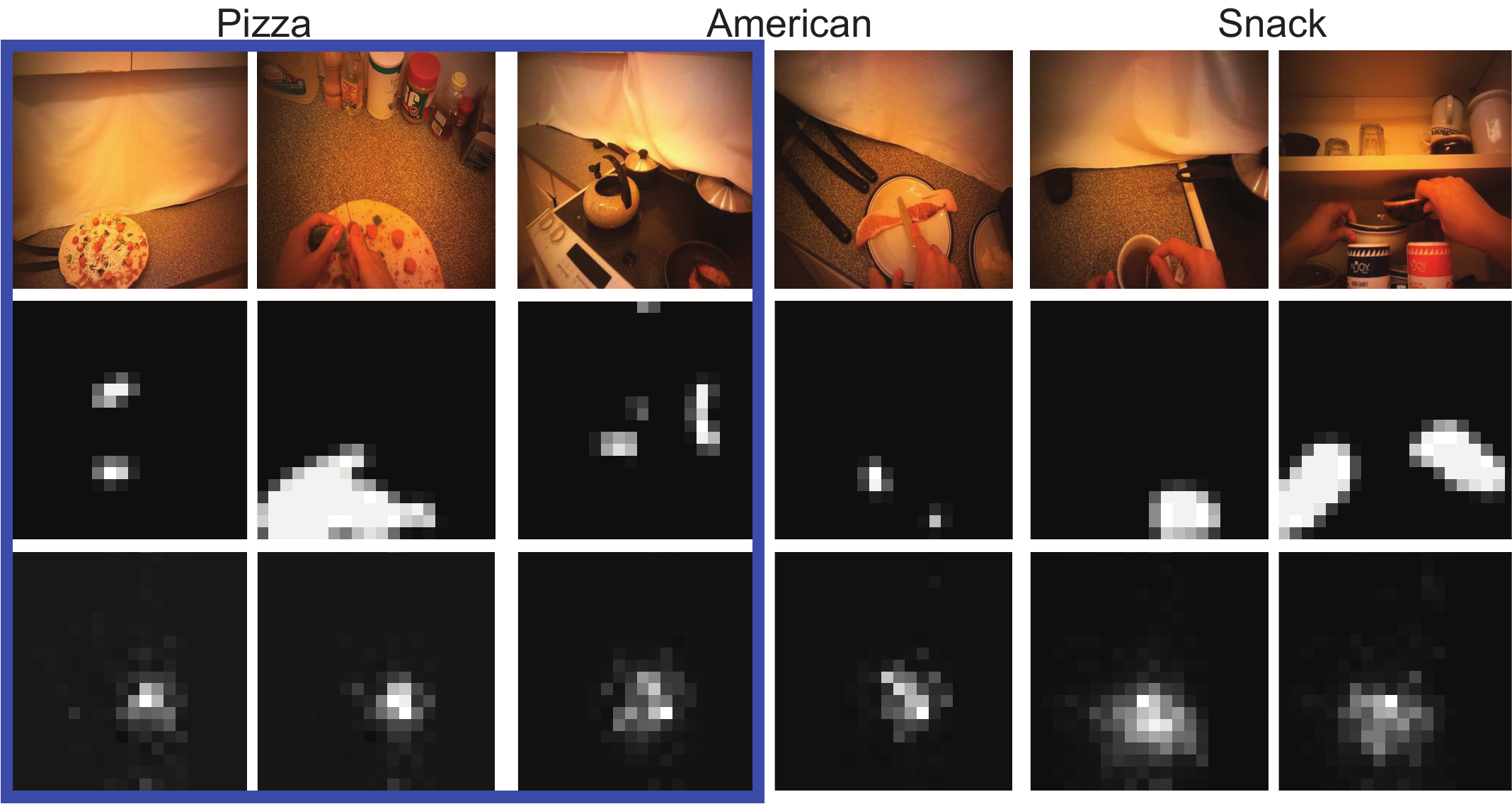}
  \caption{Sample frames from GTEA gaze+ dataset and hand segmentations using DeepLab~\cite{chen14semantic} and our learned manipulation maps. Three wrong predictions are shown with the blue box.}
  \label{fig:handSegSample}
\end{figure}

\paragraph{Manipulation point (MP)} 
To answer whether manipulation cues (during reaching, moving, or grasping objects) can improve gaze prediction, we conducted a behavioral experiment. That is, over the first 4 videos of the GTEA gaze dataset (i.e., vid1, vid2, vid3, and vid5; sampled every $10$ frames), 5 subjects were asked to click where they thought the egocentric viewer has been looking while manipulating objects.
The 2nd row of Fig.~\ref{fig:clicks} is showing the clicked points by subjects.
The average pairwise correlation among subjects, in terms of their clicked locations, over four videos in order are: $0.626, 0.678, 0.613,$ and $0.775$. The high correlation values indicate strong agreement among subjects in predicting gaze. 

We checked the agreement of manipulation map locations with fixation locations. To this end, we computed a smoothed clicked map by
convolving the binary click map on each frame, made of 5 clicks, with a small Gaussian and generated a heatmap.
We, then, computed the NSS over 4 videos. The results in order are: $1.90, 3.82, 2.4,$ and $2.22$ (mean=$2.6$). 
It means that subjects are much better than chance and any of the other models in guessing fixation locations conditioned on where the hands touch the object. 

We further looked into the role of hands in conjunction with manipulation point.
For this purpose, we classified frames into 3 categories: 
\textit{no-hands}, \textit{one-hand}, and \textit{two-hands} and measured NSS values over each category. 
The results for each video and frame category is summarized in the 1st row of Fig.~\ref{fig:clicks}. 
Mean NSS scores over 4 videos for 3 cases in order are: $1.82, 2.89$, and $3.08$.
Results demonstrate that subjects did the best when both hands were visible. 
The high correlation among clicks and fixations when hands are visible indicates that hands can be strong cues for predicting where one may look.

To see if manipulation points can further contribute to models,
we asked a new subject to guess gaze locations over all test frames of the GTEA dataset.
We then built a manipulation point map and an MP-augmented map by adding the MP map to the prediction maps of the
recurrent model. Results are shown in Table~\ref{tab:GTEA} using the recurrent model with VGG16 features.
We find that (1) the manipulation point map does better than any of
the spatial biases and the recurrent model in terms of NSS, and (2) the recurrent model augmented with manipulation map
performs better than the original recurrent model. 
This further corroborates the fact that manipulation points are strong features 
for predicting where a person may look during daily tasks.

\begin{table}[t]
\caption{Accuracy of deep prediction maps augmented with manipulation point (MP) maps over the GTEA Gaze dataset.
Augmenting Recurrent model (VGG16 features) with MP ($4$-th row) has AUCs higher than $0.878$ reported for~\cite{li2013learning}.}
\label{tab:GTEA}
\begin{center}\mbox{
\renewcommand{\tabcolsep}{0.8mm}
\begin{small}
\begin{tabular}{l|c|ccc|cc|c}
Metric & Augment & GRU (VGG16) & AFM & Gauss & MP \\
  \hline
  \hline
\multirow{2}{*}{NSS}& w/o MP& \textbf{1.542} &  1.470 & 0.950 & 2.207   \\
& w MP & \textbf{2.293} & 2.310 & 2.171 & -   \\
 \hline
\multirow{2}{*}{AUC}& w/o MP & \textbf{0.856} &   0.836  & 0.748  & 0.802 \\
& w MP&  \textbf{0.887} &0.871 &     0.858 &  -   \\
\end{tabular}
\end{small}
}\end{center}
\end{table}

\subsection{Cue Combination}  
Finally, all attention maps including Recurrent model (VGG16), All-BU, vanishing points, and learned manipulation maps (MP) are combined via regression. 
Results are shown in Table~\ref{tab:last}.
This final model does better than other models on both databases, i.e., USC database (3DDS and HDB sequences) and GTEA Gaze+ (Pizza, American, and Snack sequences). 
Obviously, the combination of All-Bottom-up model (combination of several bottom-up models) does not show a significantly better
result as the nature of the databases are task-specific and such models lack information regarding top-down attentional cues.
In comparison to Li \etal~\cite{li2013learning}, the deep recurrent model and deep regression models achieve a similar performance, indicating that the deep features are powerful enough to achieve a performance as good as a relatively complicated probabilistic model in combination of several different feature cues. 

The proposed integrated model outperforms all the models and improves over Li \etal~\cite{li2013learning} with accuracy of 0.90 versus 0.87.
This is consistent with the fact that manipulation point is a strong predictor of gaze location in egocentric vision
and potentially several top-down and task specific cues are playing a major role in egocentric gaze prediction. 

\begin{table}[t]
\caption{Accuracy of the final combined model. On USC videos (3DDS and HDB), the model includes, vanishing points (only 3DDS), All-BU and recurrent deep model. On GTEA gaze+, All-BU, recurrent deep model and learned manipulation maps are combined. 
The deep recurrent model, deep regression model, combination of all BU models are also reported.
For GTEA gaze+, the mean performance (Pizza, American, and Snack sequences) and the results of Li \etal~\cite{li2013learning} are provided for better
comparison. The human upper-bounds are AUC=0.97 and NSS=19.95, showing a significant gap between human and machine.}
\label{tab:last}
\begin{center}
\begin{footnotesize}
\renewcommand{\tabcolsep}{0.7mm}
\begin{tabular}{l|c|cc|ccc|c}
Model & Score  & 3DDS & HDB & Pizza & Amer. & Snack & GTEA Gaze+\\
  \hline
  \hline 
Integrated & NSS & 1.797 & 2.09 & 2.258 & 2.296 & 2.271 & \textbf{2.275} \\
model & AUC & 0.82 & 0.79 & 0.91 & 0.90 & 0.90 & \textbf{0.90}\\
\hline
Deep & NSS & 1.530 & 1.798 & 1.696 & 1.709 & 1.984 & 1.796 \\
recurrent & AUC & 0.82 & 0.82 & 0.87 & 0.86 & 0.89 & 0.87\\
\hline
Deep & NSS & 1.588 & 2.11 & 1.640 & 2.045 & 2.011 & 1.899 \\
regression & AUC & 0.80 & 0.75 & 0.88 & 0.88 & 0.88 & 0.88 \\
\hline
All & NSS & 1.112 & 0.902 & 1.064 & 1.101 & 1.176 & 1.114  \\
bottom-up & AUC & 0.77 & 0.77 & 0.78 & 0.77 & 0.78 & 0.78 \\
\hline
Average & NSS & 1.588 & 2.052 & 1.682  & 1.986 & 2.040 & {1.902}  \\
Fixation map & AUC & 0.80 & 0.74 & 0.89 & 0.89 & 0.88 & {0.89} \\
\hline
Li \etal~\cite{li2013learning} & AUC & ---  & ---  & ---  &  --- &  --- & 0.87\\
\end{tabular}
\end{footnotesize}
\end{center}
\end{table}

\subsection{Number of Subjects and Frames} 
\begin{figure}[t]
  \centering
    \includegraphics[width=8.3cm,height=2.3cm]{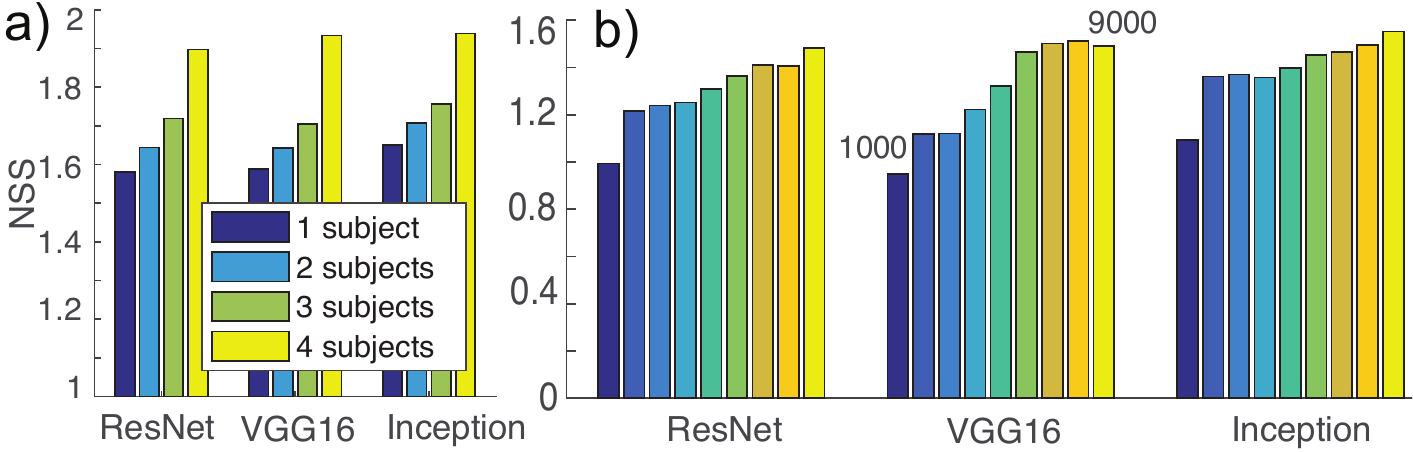}
  \caption{The effect of the number of subjects and frames}
  \label{fig:subj}  
\end{figure}

To understand the effect of the number of subjects on the recurrent model, 
we trained the recurrent deep model from data of $m$ subjects and tested it over the remaining subjects (using all combinations i.e., ${m}\choose{i}$, $i=1..4$) for each video sequence. Fig.~\ref{fig:subj}.a shows average scores over 5 videos of USC and GTEA Gaze+. As it is depicted, increasing the number of subjects improves the performance.

We further looked into the effect of the number of frames. To this end, we increased the number of training frames in steps of $1000$ 
(selected from each train video; several runs) and trained a model over each video. The learned model was then applied to the whole test video. 
The results are summarized in Fig.~\ref{fig:subj}.b. As depicted, higher number of training frames results in better accuracy.

\subsection{Knowledge Transfer} 
To assess the generalization power of the proposed recurrent deep model, we trained the model using all the data of one video and applied it to all the data of another video. We repeated the task across databases. In other words, we trained on a sequence from USC database (3DDS and HDB sequences) and employed the model to a sequence from GTEA Gaze+ (Pizza, American, and Snack) and vice versa.

\begin{figure}[htbp]
  \centering
    \includegraphics[width=8.2cm,height=3.2cm]{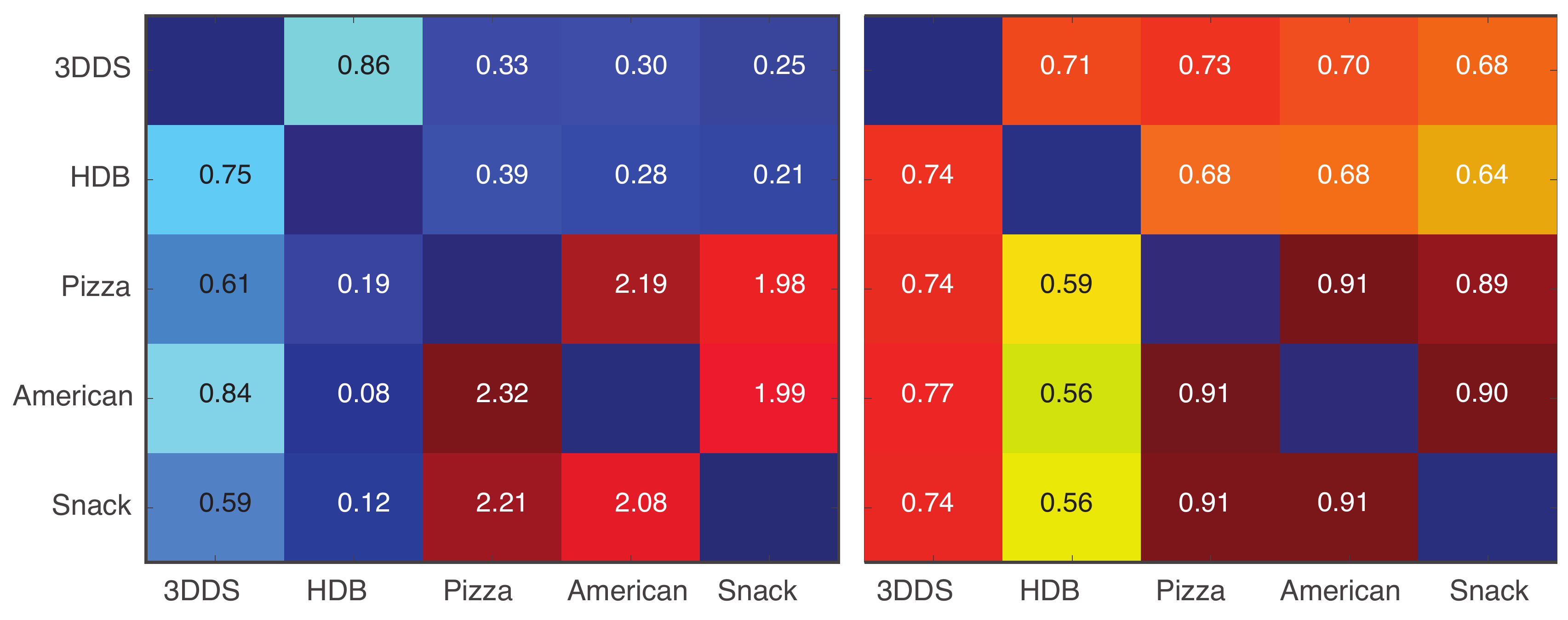}
  \caption{Confusion matrices of applying a model trained over one task to another using VGG16 features (left NSS, right AUC) }
  \label{fig:CM}  
\end{figure}

The results are summarized in Fig.~\ref{fig:CM} in terms of a confusion matrix of NSS and AUC scores. 
As depicted, all the results are above chance using both NSS and AUC scores. 
The confusion matrices show a cluster around Pizza, Snack and American video sequences (GTEA Gaze+)
indicating higher similarity among them. This is not surprising as they have been following a similar task (cooking)
in a similar environment (kitchen) where just a different meal is prepared.

To the contrary, the models trained on the HDB sequence generalize least to other sequences.
A possible reason could be that the HDB task is significantly different than the tasks of other sequences.
An example of each task in the corresponding sequences provided in Fig.~\ref{fig:MapSamples}.
Further, it has fixed background, small center-bias, and no self-motion, 

This experiment shows that it is possible to learn a model and successfully apply it to tasks that 
have a generic similarity, e.g., cooking (American, Snack, and Pizza sequences
are all involved with cooking different recipes).

\subsection{Activity and Task Prediction} 
\begin{figure}[t]
  \centering
    \includegraphics[width=8.2cm,height=3.2cm]{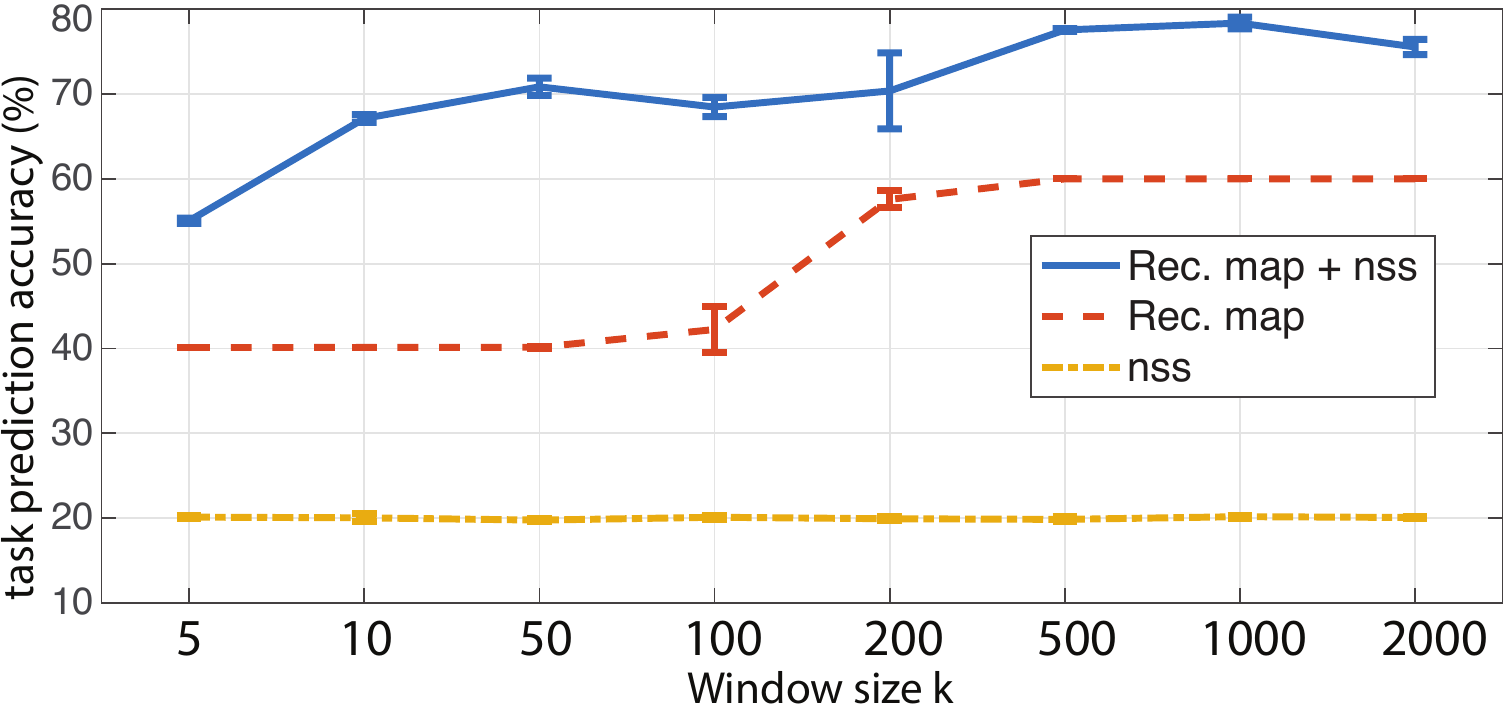}
  \caption{Task prediction over USC and GTEA Gaze+ videos}
  \label{fig:taskPred}  
\end{figure}

Facilitating task and activity recognition is one of the main motivations behind gaze prediction in egocentric vision. We, thus, investigated the use of the recurrent model for activity and task prediction. 
For this purpose, we chose 2000 windows of frames, each of size $k$, from each train video, randomly.
 Three types of features were computed, including, (a) \textit{average maps of the recurrent model with VGG16 over the window}, (b) \textit{concatenation of NSS values over frames} (a $k$D vector), and (c) \textit{augmentation of a \& b}. 
 The latter is motivated by the fact that stimulus plus behavior improves the result since behavior (here gaze) 
offers additional information regarding the task. 
We trained a linear SVM to map features to video labels (1 out of 5) and plotted the performance as a function of window size. Fig.~\ref{fig:taskPred} shows the accuracies over 2000 test windows from each test video. 
The results show that decoding activity using only NSS vector is about chance. 
The recurrent model, however, does well specially for larger windows where the average map approaches the AFM of each video. 

Augmenting the predictions with NSS values (case $c$) corresponds to the best results for
activity prediction. Notice that subjects have different gaze behaviors while executing 
tasks. Thus, the average is obtained with respect to the prediction and each subject's NSS value.
This analysis shows that gaze can be used to further improve activity and 
tasks recognition and has applications in egocentric vision.

\section{Discussion and Conclusion}

We learn the following lessons from our investigation.

a) Gaze control is based on prediction and hence inherently tied to task- and goal-relevant regions and the relative knowledge of the observer about the scene.

This is even more the case in egocentric vision, where the subject needs visual information to coordinate her manual actions in dexterous and task-effective ways. 

b) The central spatial bias is due to the tendency we have to align head and eyes with our hands when manipulating something. Due to our biomechanics we also tend to use our visual bottom hemifield more than the top one. Spatial biases are indeed completely uninformed by content, but perform well in predicting egocentric gaze.

c) Bottom up saliency is of limited relevance in strongly task-driven eye movements.
Moreover considering the optical flow feature, external motion is certainly a salient feature, but in egocentric vision indeed motion is mostly self-determined and likely ignored by the subject.

d) Hands are hardly a predictor of gaze, if not for their vicinity to the objects of interest, indeed hands are barely looked at during object manipulation and are under tightly linked to fixations which in turn are under top-down control.
Similarly, manipulation points are targeted when a grasp is planned and started but once the object is in the hand the gaze moves on to where the action is going to take place.

e) The deep learning model implicitly learns relevant features, which are not just the hands or the fingers but also object affordances or task-specific characteristics. 

Based on these findings, we foresee several future steps, including,
   (1) further investigation of top-down factors, in specific manipulation point, (2) building egocentric vision databases with a diverse set of
   stimuli and larger number of subjects, and (3) studying the robustness of
   algorithms.

{\small
$\times$Authors thank NVIDIA for the GPUs used in this work.$\times$}

\balance
{\small
\bibliographystyle{ieee}
\bibliography{iccv2016}

\begin{thebibliography}{10}\itemsep=-1pt

\bibitem{tensorflow2015-whitepaper}
M.~Abadi, A.~Agarwal, P.~Barham, E.~Brevdo, Z.~Chen, C.~Citro, G.~S. Corrado,
  A.~Davis, J.~Dean, M.~Devin, S.~Ghemawat, I.~Goodfellow, A.~Harp, G.~Irving,
  M.~Isard, Y.~Jia, R.~Jozefowicz, L.~Kaiser, M.~Kudlur, J.~Levenberg,
  D.~Man\'{e}, R.~Monga, S.~Moore, D.~Murray, C.~Olah, M.~Schuster, J.~Shlens,
  B.~Steiner, I.~Sutskever, K.~Talwar, P.~Tucker, V.~Vanhoucke, V.~Vasudevan,
  F.~Vi\'{e}gas, O.~Vinyals, P.~Warden, M.~Wattenberg, M.~Wicke, Y.~Yu, and
  X.~Zheng.
\newblock {TensorFlow}: Large-scale machine learning on heterogeneous systems,
  2015.
\newblock Software available from tensorflow.org.

\bibitem{bak2016two}
{\c{C}}.~Bak, A.~Erdem, and E.~Erdem.
\newblock Two-stream convolutional networks for dynamic saliency prediction.
\newblock {\em arXiv}, 2016.

\bibitem{bazzani2016recurrent}
L.~Bazzani, H.~Larochelle, and L.~Torresani.
\newblock Recurrent mixture density network for spatiotemporal visual
  attention.
\newblock {\em arXiv}, 2016.

\bibitem{belardinelli2016s}
A.~Belardinelli, M.~Y. Stepper, and M.~V. Butz.
\newblock It's in the eyes: Planning precise manual actions before execution.
\newblock {\em Journal of vision}, 16(1):18--18, 2016.

\bibitem{bertasius2016first}
G.~Bertasius, H.~S. Park, S.~X. Yu, and J.~Shi.
\newblock First person action-object detection with egonet.
\newblock {\em arXiv}, 2016.

\bibitem{betancourt2015evolution}
A.~Betancourt, P.~Morerio, C.~S. Regazzoni, and M.~Rauterberg.
\newblock The evolution of first person vision methods: A survey.
\newblock {\em IEEE Transactions on Circuits and Systems for Video Technology},
  2015.

\bibitem{borji2016vanishing}
A.~Borji.
\newblock Vanishing point detection with convolutional neural networks.
\newblock {\em arXiv preprint arXiv:1609.00967}, 2016.

\bibitem{borji2013state}
A.~Borji and L.~Itti.
\newblock State-of-the-art in visual attention modeling.
\newblock {\em IEEE Trans. Pattern Analysis and Machine Intelligence}, 2013.

\bibitem{borji2012probabilistic}
A.~Borji, D.~N. Sihite, and L.~Itti.
\newblock Probabilistic learning of task-specific visual attention.
\newblock In {\em CVPR}, pages 470--477. IEEE, 2012.

\bibitem{borji2013quantitative}
A.~Borji, D.~N. Sihite, and L.~Itti.
\newblock Quantitative analysis of human-model agreement in visual saliency
  modeling: a comparative study.
\newblock {\em IEEE Transactions on Image Processing}, 22(1):55--69, 2013.

\bibitem{vondrick2016nips}
V.~Carl, P.~Hamed, and T.~Antonio.
\newblock Generating videos with scene dynamics.
\newblock In {\em Advances In Neural Information Processing Systems}, pages
  613--621. NIPS, 2016.

\bibitem{chen14semantic}
L.-C. Chen, G.~Papandreou, I.~Kokkinos, K.~Murphy, and A.~L. Yuille.
\newblock Semantic image segmentation with deep convolutional nets and fully
  connected crfs.
\newblock In {\em ICLR}, 2015.

\bibitem{chung2014empirical}
J.~Chung, C.~Gulcehre, K.~Cho, and Y.~Bengio.
\newblock Empirical evaluation of gated recurrent neural networks on sequence
  modeling.
\newblock {\em arXiv preprint arXiv:1412.3555}, 2014.

\bibitem{deng2009imagenet}
J.~Deng, W.~Dong, R.~Socher, L.-J. Li, K.~Li, and L.~Fei-Fei.
\newblock Imagenet: A large-scale hierarchical image database.
\newblock In {\em CVPR}, pages 248--255. IEEE, 2009.

\bibitem{fathi2012learning}
A.~Fathi, Y.~Li, and J.~M. Rehg.
\newblock Learning to recognize daily actions using gaze.
\newblock In {\em ECCV}, pages 314--327. Springer, 2012.

\bibitem{ghosh2012discovering}
J.~Ghosh, Y.~J. Lee, and K.~Grauman.
\newblock Discovering important people and objects for egocentric video
  summarization.
\newblock In {\em CVPR}, pages 1346--1353. IEEE, 2012.

\bibitem{guo2008spatio}
C.~Guo, Q.~Ma, and L.~Zhang.
\newblock Spatio-temporal saliency detection using phase spectrum of quaternion
  fourier transform.
\newblock In {\em CVPR}, pages 1--8. IEEE, 2008.

\bibitem{hadsell2006dimensionality}
R.~Hadsell, S.~Chopra, and Y.~LeCun.
\newblock Dimensionality reduction by learning an invariant mapping.
\newblock In {\em CVPR}, volume~2, pages 1735--1742. IEEE, 2006.

\bibitem{harel2006graph}
J.~Harel, C.~Koch, and P.~Perona.
\newblock Graph-based visual saliency.
\newblock In {\em NIPS}, pages 545--552, 2006.

\bibitem{horn1981determining}
B.~K. Horn and B.~G. Schunck.
\newblock Determining optical flow.
\newblock {\em Artificial intelligence}, 17(1-3):185--203, 1981.

\bibitem{hou2007saliency}
X.~Hou and L.~Zhang.
\newblock Saliency detection: A spectral residual approach.
\newblock In {\em CVPR}, pages 1--8. IEEE, 2007.

\bibitem{huang2015salicon}
X.~Huang, C.~Shen, X.~Boix, and Q.~Zhao.
\newblock Salicon: Reducing the semantic gap in saliency prediction by adapting
  deep neural networks.
\newblock In {\em ICCV}, pages 262--270, 2015.

\bibitem{huang2018predicting}
Y.~Huang, M.~Cai, Z.~Li, and Y.~Sato.
\newblock Predicting gaze in egocentric video by learning task-dependent
  attention transition.
\newblock {\em arXiv preprint arXiv:1803.09125}, 2018.

\bibitem{itti2005bayesian}
L.~Itti and P.~F. Baldi.
\newblock Bayesian surprise attracts human attention.
\newblock In {\em NIPS}, pages 547--554, 2005.

\bibitem{itti1998model}
L.~Itti, C.~Koch, and E.~Niebur.
\newblock A model of saliency-based visual attention for rapid scene analysis.
\newblock {\em IEEE PAMI}, 1998.

\bibitem{jain2015gaze}
E.~Jain, Y.~Sheikh, A.~Shamir, and J.~Hodgins.
\newblock Gaze-driven video re-editing.
\newblock {\em ACM Transactions on Graphics (TOG)}, 34(2):21, 2015.

\bibitem{kingma2014adam}
D.~Kingma and J.~Ba.
\newblock Adam: A method for stochastic optimization.
\newblock {\em arXiv preprint arXiv:1412.6980}, 2014.

\bibitem{land2001ways}
M.~F. Land and M.~Hayhoe.
\newblock In what ways do eye movements contribute to everyday activities?
\newblock {\em Vision research}, 41(25), 2001.

\bibitem{li2013learning}
Y.~Li, A.~Fathi, and J.~Rehg.
\newblock Learning to predict gaze in egocentric video.
\newblock In {\em ICCV}, pages 3216--3223, 2013.

\bibitem{matsuo2014attention}
K.~Matsuo, K.~Yamada, S.~Ueno, and S.~Naito.
\newblock An attention-based activity recognition for egocentric video.
\newblock In {\em CVPR Workshops}, pages 551--556, 2014.

\bibitem{nakashima2015saliency}
R.~Nakashima, Y.~Fang, Y.~Hatori, A.~Hiratani, K.~Matsumiya, I.~Kuriki, and
  S.~Shioiri.
\newblock Saliency-based gaze prediction based on head direction.
\newblock {\em Vision research}, 117:59--66, 2015.

\bibitem{navalpakkam2006integrated}
V.~Navalpakkam and L.~Itti.
\newblock An integrated model of top-down and bottom-up attention for
  optimizing detection speed.
\newblock In {\em CVPR}, volume~2, pages 2049--2056. IEEE, 2006.

\bibitem{park20123d}
H.~S. Park, E.~Jain, and Y.~Sheikh.
\newblock 3d social saliency from head-mounted cameras.
\newblock In {\em NIPS}, pages 431--439, 2012.

\bibitem{peters2007beyond}
R.~J. Peters and L.~Itti.
\newblock Beyond bottom-up: Incorporating task-dependent influences into a
  computational model of spatial attention.
\newblock In {\em CVPR}, pages 1--8. IEEE, 2007.

\bibitem{pirsiavash2012detecting}
H.~Pirsiavash and D.~Ramanan.
\newblock Detecting activities of daily living in first-person camera views.
\newblock In {\em CVPR 2012}, pages 2847--2854. IEEE, 2012.

\bibitem{polatseknovelty}
P.~Polatsek, W.~Benesova, L.~Paletta, and R.~Perko.
\newblock Novelty-based spatiotemporal saliency detection for prediction of
  gaze in egocentric video.

\bibitem{ren2009egocentric}
X.~Ren and M.~Philipose.
\newblock Egocentric recognition of handled objects: Benchmark and analysis.
\newblock In {\em CVPR Workshops}, pages 1--8. IEEE, 2009.

\bibitem{rudoy2013learning}
D.~Rudoy, D.~B. Goldman, E.~Shechtman, and L.~Zelnik-Manor.
\newblock Learning video saliency from human gaze using candidate selection.
\newblock In {\em CVPR}, pages 1147--1154, 2013.

\bibitem{seo2009static}
H.~J. Seo and P.~Milanfar.
\newblock Static and space-time visual saliency detection by self-resemblance.
\newblock {\em Journal of vision}, 9(12):15--15, 2009.

\bibitem{simonyan2014two}
K.~Simonyan and A.~Zisserman.
\newblock Two-stream convolutional networks for action recognition in videos.
\newblock In {\em NIPS}, pages 568--576, 2014.

\bibitem{sprague2003eye}
N.~Sprague and D.~Ballard.
\newblock Eye movements for reward maximization.
\newblock In {\em NIPS}, page None, 2003.

\bibitem{su2016detecting}
Y.-C. Su and K.~Grauman.
\newblock Detecting engagement in egocentric video.
\newblock {\em arXiv}, 2016.

\bibitem{fpvvstpv}
C.~Tan, H.~Goh, V.~Chandrasekhar, L.~Li, and J.~Lim.
\newblock Understanding the nature of first-person videos: Characterization and
  classification using low-level features.
\newblock 2014.

\bibitem{tatler2007central}
B.~W. Tatler.
\newblock The central fixation bias in scene viewing: Selecting an optimal
  viewing position independently of motor biases and image feature
  distributions.
\newblock {\em Journal of Vision}, 7(14):4--4, 2007.

\bibitem{Xu_2015_CVPR}
J.~Xu, L.~Mukherjee, Y.~Li, J.~Warner, J.~M. Rehg, and V.~Singh.
\newblock Gaze-enabled egocentric video summarization via constrained
  submodular maximization.
\newblock In {\em CVPR}, June 2015.

\bibitem{yamada2010can}
K.~Yamada, Y.~Sugano, T.~Okabe, Y.~Sato, A.~Sugimoto, and K.~Hiraki.
\newblock Can saliency map models predict human egocentric visual attention?
\newblock In {\em ACCV Workshops}, pages 420--429. Springer, 2010.

\bibitem{yonetani2016visual}
R.~Yonetani, K.~M. Kitani, and Y.~Sato.
\newblock Visual motif discovery via first-person vision.
\newblock In {\em ECCV}, pages 187--203. Springer, 2016.

\bibitem{yu2013joint}
C.~Yu and L.~B. Smith.
\newblock Joint attention without gaze following: Human infants and their
  parents coordinate visual attention to objects through eye-hand coordination.
\newblock {\em PloS one}, 8(11):e79659, 2013.

\bibitem{Zhang2017}
M.~Zhang, K.~T. Ma, J.~H. Lim, Q.~Zhao, and J.~Feng.
\newblock Deep future gaze: Gaze anticipation on egocentric videos using
  adversarial networks.
\newblock In {\em CVPR}, 2017.

\end{thebibliography}
}

\end{document}